# Using image-extracted features to determine heart rate and blink duration for driver sleepiness detection

Erfan Darzi,[1] Armin Mohammadie-Zand,[2] Hamid Soltanian-Zadeh[1,3]

[1] Control and Intelligence Processing Center of Excellence (CIPCE), School of Electrical and Computer Engineering, College of Engineering, University of Tehran, Tehran, Iran
[2] Department of Electerical Engineering,Amirkabir University of Technology, Tehran, Iran
[3] Medical Image Analysis Laboratory, Henry Ford Health System, Detroit, MI, USA
erfandarzi@ut.ac.ir, hszadeh@ut.ac.ir, hsoltan1@hfhs.org

*Abstract*—Heart rate and blink duration are two vital physiological signals which give information about cardiac activity and consciousness. Monitoring these two signals is crucial for various applications such as driver drowsiness detection. As there are several problems posed by the conventional systems to be used for continuous, long-term monitoring, a remote blink and ECG monitoring system can be used as an alternative. For estimating the blink duration, two strategies are used. In the first approach, pictures of open and closed eyes are fed into an Artificial Neural Network (ANN) to decide whether the eyes are open or close. In the second approach, they are classified and labeled using Linear Discriminant Analysis (LDA). The labeled images are then be used to determine the blink duration. For heart rate variability, two strategies are used to evaluate the passing blood volume: Independent Component Analysis (ICA); and a chrominance based method. Eye recognition yielded 78-92% accuracy in classifying open/closed eyes with ANN and 71-91% accuracy with LDA. Heart rate evaluations had a mean loss of around 16 Beats Per Minute (BPM) for the ICA strategy and 13 BPM for the chrominance based technique.

*Keywords—Blink duration, Linear discriminant analysis, Artificial Neural Network (ANN), Heart rate estimation*

## I. INTRODUCTION

Driving while tired is similar to driving while drunk in many aspects, as tired drivers display behaviors identical to drunk drivers: delay in response times, failure to think, and inadequate attention to the environment. The issue is also worsened as it may be hard for drivers to perceive indications of drowsiness. A study about this issue by the AAA Foundation for Traffic Safety found that 21% of deadly crashes included a sleepy driver [1]. According to another study, NHTSA General Estimates System (GES) statistics for a four-year span indicates that there were an estimated 56,000 crashes annually in which driver drowsiness/fatigue was cited on the Police Accident Report (PAR) [2].

There are a few physiological signs that can be used to decide whether a driver is tired. One such sign is having difficulty in keeping eyes open. By following whether the eyes are open or closed, the blink duration can be found, and then utilized to decide how much hardship the driver has to remain wakeful. Another sign is heart rate variability, which is the change in the time between heart pulses. Sleepy and non-sleepy individuals display different heart rate variabilities [3]. Given this, our objective is to make a blink and heart rate detector using a simple RGB camera like those found in mobile phones and laptops, and under the actual-world circumstance. The output data from heart rate and blink detector could then be utilized to determine how sleepy an individual is [4].

There have been some works done in estimating an individual's heart rate and look, employing simple cameras like the ones accessible on most present-day cellphones and laptops [5], [6]. While some of these studies did not investigate sleepiness identification, their techniques may be helpful for sleepiness discovery. Heart rate detection utilizing video of the frontal face has been practiced through an assortment of strategies [7], [8]. Poh et al. [9] detected RMSE for

HR prediction from 43.76 beats per minute (BPM). They suggested considering the 60% of the width of the face and applying a blind source separator for heart rate detection to get the best results. Monkaresi et al. reached an RMSE of 8.49 BPM for HR [7], however, their prediction was under controlled conditions. Using this algorithm to identify heart pulse, the heart rate variability may be determined.

Similarly, there has been a wide range of techniques developed for eye motion tracking employing frontal face image or video [10], [11]. Chuang-Wen et al. reached 83% accuracy with dual camera smartphone in a blink detection task [12]. Since we are not keen on accurate eye segmentation, the basic eye segmentation techniques found in the literature may be used in our work. In fact, we use them in the first stage of our algorithm to obtain segmented eyes and then decide whether they are open or closed.

## II. METHODS AND MATERIALS

### A. Preprocessing

The majority of our algorithms work instantaneously and at a frame by frame processing for a video either from a webcam or a video document. For face recognition, we have used the OpenCV toolkit for face recognition [13]. This toolkit includes a variety of pre-trained functions to draw bounding boxes for the eyes. For heart rate calculation, there were two regions of interest: facial skin; and cheeks, which were obtained and pre-processed. The processing of facial skin was done by transforming the picture into the HSI color space.

### B. Eye Detection

*1) Training:*

*a) Linear Discriminant Analysis:* It was crucial for determining the blink rate and length to decide whether eyes were open or closed. Based on the twofold concept of the issue and the distinct contrast between open and closed eyes, Linear Discriminant Analysis (LDA) was used due to its capability of solving binary classification problems.

*b) Artificial Neural Network (ANN):* An ANN was also trained using the open/closed training images. Since the capability of neural networks is proven in such classification tasks, applying them seems to be a proper strategy, which would be examined in succeeding sections.

*c) Data:* The training information comprised of 100 pictures of closed eyes and 100 images of open eyes. To gather the information, we took 20 photos at once, at a sampling rate controlled by the frame rate of the webcam. This process was repeated multiple times for each class at various positions, with different brightness and distances from the camera.

*d) Preprocessing:* Images varied a lot in illumination and contrast, which might affect the performance of the classifier. To handle this issue, we histogram equalized the images before classifying them.

*e) Application of LDA:* The first 100 eigenimages were obtained using Pentland algorithm [14]. Then, the first principal component (eigenimage) was used for investigation. Eigenimages are the principal components of distribution of faces, or equivalently, the eigenvectors of the covariance matrix of the set of face images. LDA was then utilized to elicit the discriminant image from the eigenimages. The discriminant image displays the decision surface between the two classes. Once the discriminant image was acquired, the image to be tested was projected onto the fisher face and labeled closed or open using an appropriate threshold.

*2) Testing:* Two hundred consecutive frames were acquired to test the eye classification task.

*3) Determining Blink Duration:* The blink duration and the eyes status (open or closed) were determined and displayed in the python command window in real-time. One post-processing step was used to remove false negatives, which was crucial for calculating the blink duration. For any negative frame (open eyes) detected between two positive frames (closed eyes), the negative frame was ignored. This is reasonable because a blink is expected to last more than a single frame.

### C. Heart Pulse Detection

As the heart circulates blood through the body, the volume of blood passing through the vessels changes. The change in blood volume is noticeable in skin color. By recording the RGB images of the face, the blood volume passing through the face vessels can be estimated, from which the heart pulse and other related measurements can be made. Two unique strategies were utilized to determine the heart beat: independent component analysis (ICA); and a chrominance based method. These two techniques are similar in their early steps, like defing a region of interest (ROI), separating the red, green, and blue channels, and generated a time series for each channel by averaging all pixel values in the ROI.

*1) Independent Component Analysis:* ICA is a technique that attempts to find an ideal linear transform from *n* signals into *n* statistically independent signals. Numerous papers have demonstrated that performing ICA on the three standardized color channels of an individual's skin yields a signal similar to the blood volume pulse [15]. Equation (1) demonstrates how the red channel may be standardized with the green and blue channels following a similar syste, where $\mu_r$ and $\sigma_r$ are the mean and variance of the input signal, respectively. Equation (2) shows the ICA trasfromation on information signal x(t) which utilizes the unmixing matrix W to obtain output signal s(t). Figure 1 frames what ICA actually does [16].

$$r_{norm} = \frac{r - \mu_r}{\sigma_r} \quad (1)$$

$$s(t) = Wx(t) \quad (2)$$

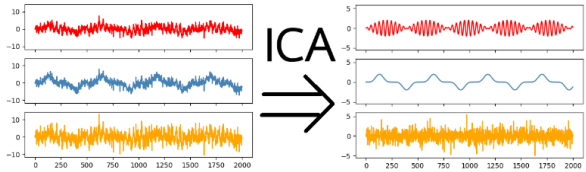

Figure 1: A schema of what ICA actually does, Color Channels Transformed into independent channels.

First, we used ICA on the standardized red, green, and blue signal to train an unmixing matrix that will be applied to every single future video and webcam film. None of the signals appeared to be practically equivalent to blood volume pulse by visual examination. Thus, we mapped the signal into the frequency domain by taking the FFT of each of the three channels and considering the dominant frequency as the channel's primary frequency. We then took the channel in which the dominant frequency was between 0.67 Hz and 3 Hz, corresponding to heart rates between 40 BPM and 180 BPM separately. We picked the signal whose dominant frequency had adequate size and whose frequency was nearest to the normal heart rate of 71 bpm determined from the ECG. The output of the matrix corresponding to this signal was the one we used to determine passing blood volume and heart rate. The heart rate was assessed by taking the FFT of the blood volume heart beat and taking the channel at the predominant frequency between 0.67 Hz and 3 Hz and after that myltiplying the rate by 60 to convert the result to bpm.

*2) Chrominance Based Method:* The most notable advantage of the chrominance based technique [8] is its robustness to movement. Two sorts of reflections are caught by a camera as light reflects off of skin: specular; and diffuse. The specular reflection is attributed as it were to the light source, and not to any physiological origins, while the diffuse reflection is light that has been dissipated from the veins, and hence contains the data expected to determine the blood volume pulse. Haan et al.[8] suggested that by accepting a normalized skin color, which they observed to be the same over all skin types, standardized RGB can be mapped into the plane orthogonal to the specular reflection component, producing two chrominance components, X and Y. These components are bandpass filtered by means of a Butterworth bandpass filter of order 3 for frequencies 0.67 - 3 Hz, and then used to determine the blood volume pulse. Equations (3)-(6) outline the calculations. R, G, and B are color components, and $X_{filtered}$, $Y_{filtered}$ represent the filtered chrominance components.

$$X = 3R - 2G \quad (3)$$

$$Y = 1.5R + G - 1.5B \quad (4)$$

$$\alpha = \frac{\sigma(X_{filtered})}{\sigma(Y_{filtered})} \quad (5)$$

$$S = X_{filtered} - \alpha Y_{filtered} \quad (6)$$

According to Haan et al.[8], heart rate is assessed by taking the FFT of the blood volume heart beat and taking the predominant frequency in the range 0.67 - 3 Hz. Figure 2 demonstrates the signal in the frequency domain. The dominant component of the blood volume heart beat signal has a frequency of 61 frames or a period of about 1 second with respect to the frame rate of 60 fps. So, it can be concluded that there would be 60 heart beat pulses in a minute.

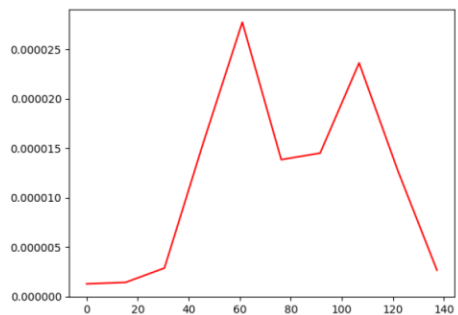

Figure 2: The signal in the frequency domain, showing a peak at 61 beats per minute.

## III. EVALUATION

### A. Eye Detection

*1) Eye Classification Training:* Figure 3 demonstrates the discriminant image, i.e., the image of the decision boundary for the LDA classifier. Images which are more similar than the discriminant image to the class 'open' are classified as open and otherwise, classified as closed.

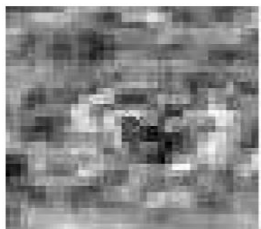

Figure 3: A discriminant image.

*2) Eye Classification Testing:* Tables I and II show sample testing results for ANN and LDA classifiers, respectively. Two hundred consecutive pictures were assessed, a total of 12 blinks happened, each lasting 2-8 frames. For the first trial, ANN had an accuracy of about 86.5%, with 173 correct classifications out of 200 test frames. Likewise, LDA was 79% correct in the same trial, with 158 correct classifications out of 200 frames. A total of 5 trials were done in the same manner, which yielded an accuracy of 78-92% for ANN and 79-91% for LDA.

TABLE I. Sample test results for blink classification of 200 video frames using ANN.

| Truth | Correctly Classified | Incorrectly classified |
|---|---|---|
| Open | 35 | 5 |
| Shut | 138 | 22 |

TABLE II. Sample test results for blink classification of 200 video frames using LDA.

| Truth | Correctly Classified | Incorrectly classified |
|---|---|---|
| Open | 35 | 5 |
| Shut | 123 | 38 |

### B. Heart Rate Detection

To assess our heart rate detection algorithm, we used the PhotoPlethysmoGram (PPG) sensor of a smartphone as the ground truth for heart beat rate. To reduce possible noises and artifacts, we calculated the average of 30 seconds of heart pulses and considered it as the ground truth. The estimated heart rate fluctuated a lot, as the relatively large standard deviation indicates. To alleviate fluctuations, we picked the median value of consecutive frames. Table III summarizes our discoveries. The mean error for the ICA method is around 16 BPM, while for the chrominance method is approximately 13 BPM. Sample outputs of the algorithm are shown in Figure 4.

TABLE III. Results of heart rate estimation from 200 video frames. Med.:Median; Std.Dev.: Standard Deviation; Avg.: Average.

| Truth | ICA | | | Chrominance | | |
|---|---|---|---|---|---|---|
| | Med. | Std.Dev. | Error | Med. | Std.Dev. | Error |
| 87 | 75 | 28.5 | 12 | 83 | 29 | 4 |
| 64 | 53 | 21 | 11 | 81 | 12 | 11 |
| 59 | 54 | 18.1 | 5 | 44 | 32.9 | 5 |
| 102 | 79 | 17 | 23 | 113 | 22 | 11 |
| 72 | 88 | 20.4 | 16 | 70 | 5 | 2 |
| 106 | 63 | 29 | 43 | 67 | 16.8 | 39 |
| 63 | 67 | 9 | 4 | 60 | 14 | 3 |
| Avg. | | | 16 | | | 13 |

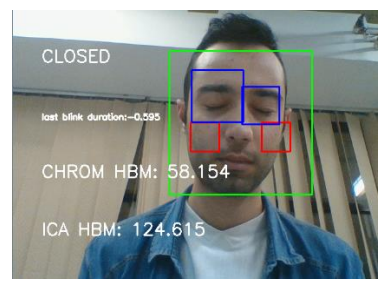

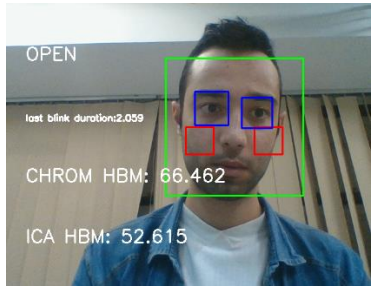

Figure 4: Sample results displaying eyes status and heart beat estimation

## IV. DISCUSSION

### A. Eye Detection

LDA testing had an accuracy of 79-91%. This indicates that the LDA algorithm considers all the training data adequately and provides a reliable beginning stage to distinguish between open and closed eyes. ANN had a slightly better performance with 78-92% accuracy. This advantage can be interpreted by the fact that ANN has numerous hyperparameters and optimizers, which all could be selected appropriately and regulated for performance maximization. It is also worth noting that the pre-trained classifiers were weak at identifying closed eyes under specific illumination conditions. This weakness arises from the constraints in the camera and the quality of the images used for the procedure, not from the algorithm itself. Because the OpenCV pre-trained classifier fails to detect closed eyes under low illumination, classifying images in which no eye has been detected as closed would improve the accuracy of the test results. However, that does not seem reasonable, since accurate eye detection is crucial for practical purposes.

### B. Heart Rate Detection

Generally, the heart rate detection algorithms presented in this paper worked better than those of similar papers [9], [15]. Although both ICA and chrominance methods attempt to improve the signal to noise ratio, these methods are not perfect. This might explain why these methods yielded inaccurate heart rate estimations. Video compression also affected the overall accuracy, which is an indispensable part of image acquisition. It seems very hard to improve the results, as we tried different lightning conditions and user positions to get better results, but the accuracy did not improve much. However, as we know that random noise error can be removed by taking an average of several trials, it could be stated that increasing the number of trials and then averaging the results may eliminate random noise and yield better results.

## V. CONCLUSION

This project was way more complicated than it looked at the first glance. The accuracy of eye identification is more than 90% under proper illumination. However, in some conditions, it is restricted by the pre-trained OpenCV classifiers for eye segmentation. More reliable strategies for segmentation could be used for future improvement. Even though we did not have much knowledge about the biophysical characteristics of the driver, we could actualize a heart rate screening and monitoring system with an improved accuracy in comparison with some previous works [15]. However, The accuracy was slightly less than some other papers [12]. Rahman et al. hit 94% performance with a 16 MP full-HD camera [17] and Chuang-Wen et al. reached 83% with dual camera smartphone in the blink detection task. In both cases, the environment was fully controlled. In spite of their strategies, this minor difference is likely because our assessment was not performed in a controlled state. For our ultimate purpose of examining driver sleepiness, the situation is not controlled, and thus, we attempted to get results that would be under the actual-world circumstance. An apparent drawback of our proposed algorithm is that the assessed heart rate fluctuated a lot. Notwithstanding, getting a rough approximation of the heart rate from a simple RGB camera is an accomplishment. Also, the fact that we could run everything real-time is an advantage that many methods are deprived of [15]. This venture is an initial move towards an appropriate sleepiness detector.